\documentclass[letterpaper,10pt,journal,twoside]{IEEEtran}
\usepackage{graphicx} 
\usepackage{multirow}
\usepackage{hyperref}
\usepackage{float}
\usepackage{amsmath}
\usepackage{booktabs}
\usepackage{multirow}
\usepackage{siunitx} 
\usepackage{xcolor}
\usepackage{colortbl}
\usepackage{tcolorbox}
\newtcbox{\revibox}{colback = violet, coltext=white, boxrule = 0pt,arc=5pt, boxsep=0pt,left=2pt,right=2pt,top=2pt, bottom=2pt, on line}
\usepackage[colorinlistoftodos]{todonotes}
\newtcbox{\reviiibox}{colback = orange, coltext=white, boxrule = 0pt,arc=5pt, boxsep=0pt,left=3pt,right=3pt,top=2pt, bottom=2pt, on line}

\IEEEoverridecommandlockouts                              

\usepackage{amssymb} 

\title{
Diff2DGS: Reliable Reconstruction of Occluded Surgical Scenes via 2D Gaussian Splatting
}
\author{Tianyi Song$^{1}$, Danail Stoyanov$^{1}$,
Evangelos Mazomenos$^{1}$, and Francisco Vasconcelos$^{1}$%
}

\begin{document}

\maketitle

\begin{abstract}
Real-time reconstruction of deformable surgical scenes is vital for advancing robotic surgery, improving intraoperative guidance, and enabling automation. Recent methods achieve dense reconstructions from da Vinci robotic surgery videos, with Gaussian Splatting (GS) offering real-time performance via graphics acceleration. However, reconstruction quality in occluded regions remains limited, and depth accuracy has not been fully assessed, as benchmarks like EndoNeRF and StereoMIS lack 3D ground truth. We propose Diff2DGS, a two-stage framework for reliable 3D reconstruction of occluded surgical scenes. First, a diffusion-based video module with temporal priors inpaints tissue occluded by instruments with high spatiotemporal consistency. Second, we adapt 2D Gaussian Splatting (2DGS) with a Learnable Deformation Model (LDM) to capture dynamic tissue deformation and anatomical geometry, and introduce adaptive depth weight to improve geometric fidelity. We further extend evaluation beyond image-quality metrics by performing quantitative depth analysis on the SCARED dataset. Diff2DGS outperforms state-of-the-art methods in both appearance and geometry, reaching 38.02 dB PSNR on EndoNeRF and 34.40 dB on StereoMIS. Our experiments also show that optimizing image quality alone does not necessarily ensure accurate 3D geometry. The code is available at \href{https://diff2dgs.github.io/}{https://diff2dgs.github.io/}.
\end{abstract}


\section{INTRODUCTION}
\subsection{Real-time perception for robot-assisted surgery}

\IEEEPARstart{I}ntraoperative 3D reconstruction is essential for robot-assisted surgery, where accurate and dynamic scene perception can support surgical navigation \cite{surgical_navigation1,surgical_navigation2}, robotic assistance \cite{robort1}, and simulation-based training \cite{train1,train2}. Early approaches relied on depth estimation \cite{depthestimation1,depthestimation2}, simultaneous localization and mapping (SLAM) \cite{slam_posenet, track2map}, and point cloud fusion \cite{pointcloudfussion}. More recent Neural Radiance Field (NeRF)-based methods \cite{nerf0,surgey_nerf0,surgey_nerf1,dynamic1} improve reconstruction fidelity by representing surgical scenes as dense neural fields, but their high computational cost limits their practicality for real-time intraoperative applications.

\begin{figure}[htbp]
  \centering
  \includegraphics[width=0.9\columnwidth]{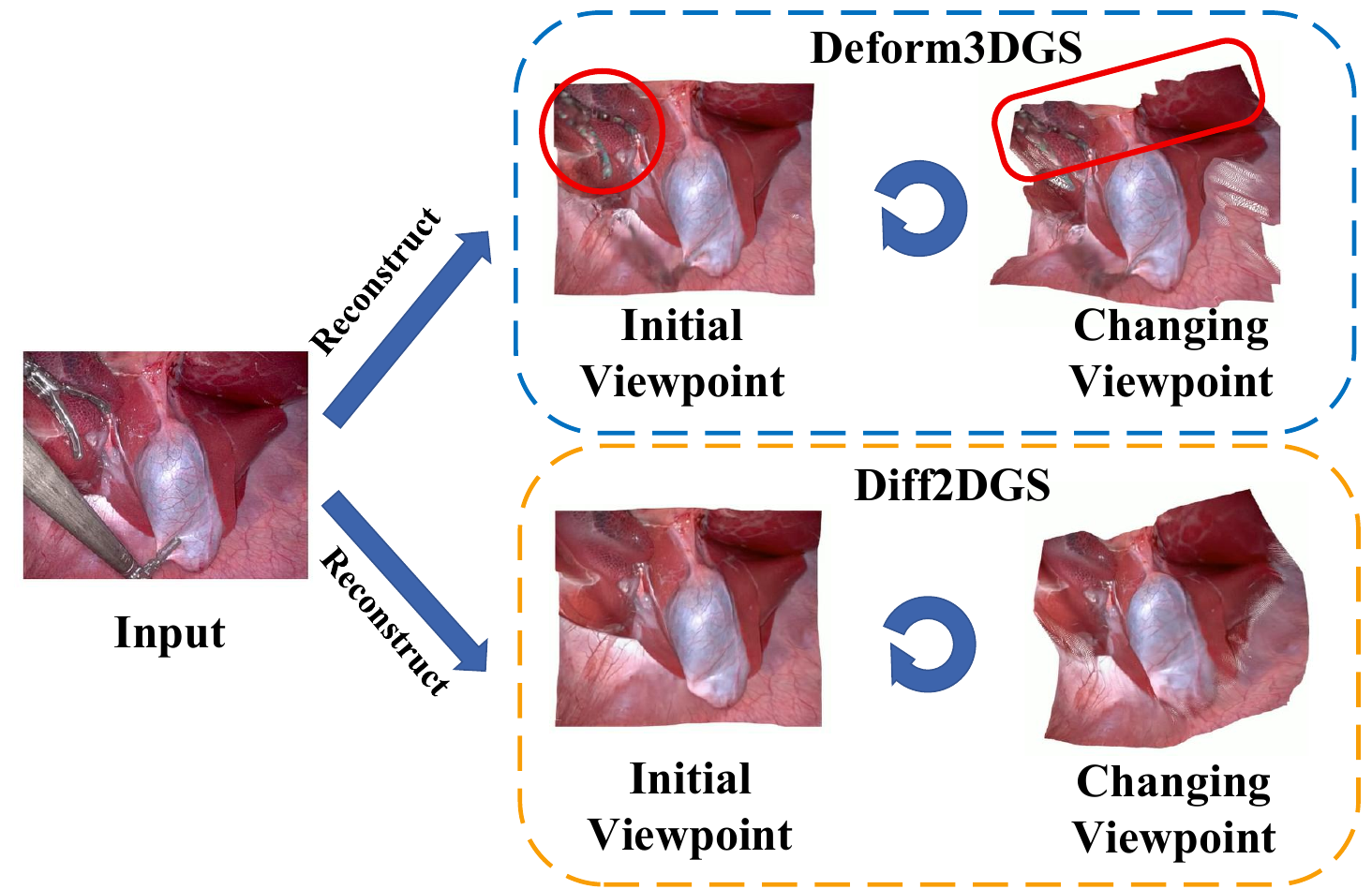}
  \caption{Traditional endoscopic scene reconstruction methods often focus primarily on image quality from the camera viewpoint, neglecting the depth accuracy of the reconstructed geometry. Consequently, when the camera viewpoint changes, reconstruction accuracy degrades significantly. We present Diff2DGS, a framework that effectively balances depth information and image quality, achieving high-quality 3D reconstruction while better eliminating artifacts in occluded regions and providing more accurate depth estimation.}
  \label{fig:single_column}
  
\end{figure}

\subsection{Dynamic surgical scene reconstruction}
3D Gaussian Splatting (3DGS) \cite{3dgs0} has recently emerged as an efficient alternative for dense scene reconstruction. Although originally designed for static scenes, it has been extended to dynamic environments through temporal deformation modeling \cite{4dgs, deformable3dgs,deformable3dgs2}.

In surgical scenarios, EndoGaussian \cite{endogaussian} and Deform3DGS \cite{deform3dgs} leverage deformation fields combined with efficient voxel encoding and lightweight deformation decoders to achieve real-time surgical scene reconstruction with dynamic tissue modeling. Endo-4DGS \cite{endo4dgs} incorporates monocular depth priors generated by Depth Anything \cite{depthanything} and introduces a confidence-guided learning strategy to mitigate uncertainties in monocular depth estimation. Similarly, Deform3DGS \cite{deform3dgs} models tissue deformation through efficient linear combination regression with learnable basis functions, enhancing representational capacity and computational efficiency. More recently, SurgicalGS~\cite{surgicalgs} has improved dynamic surgical scene reconstruction by incorporating stereo-depth priors, including multi-frame depth fusion for Gaussian initialization and depth regularization during optimization. However, existing 3DGS-based methods still struggle to accurately reconstruct deformable tissue surfaces, especially under frequent surgical instrument occlusions, which can lead to artifacts and geometric distortions.

\begin{figure*}
\includegraphics[width=\textwidth]{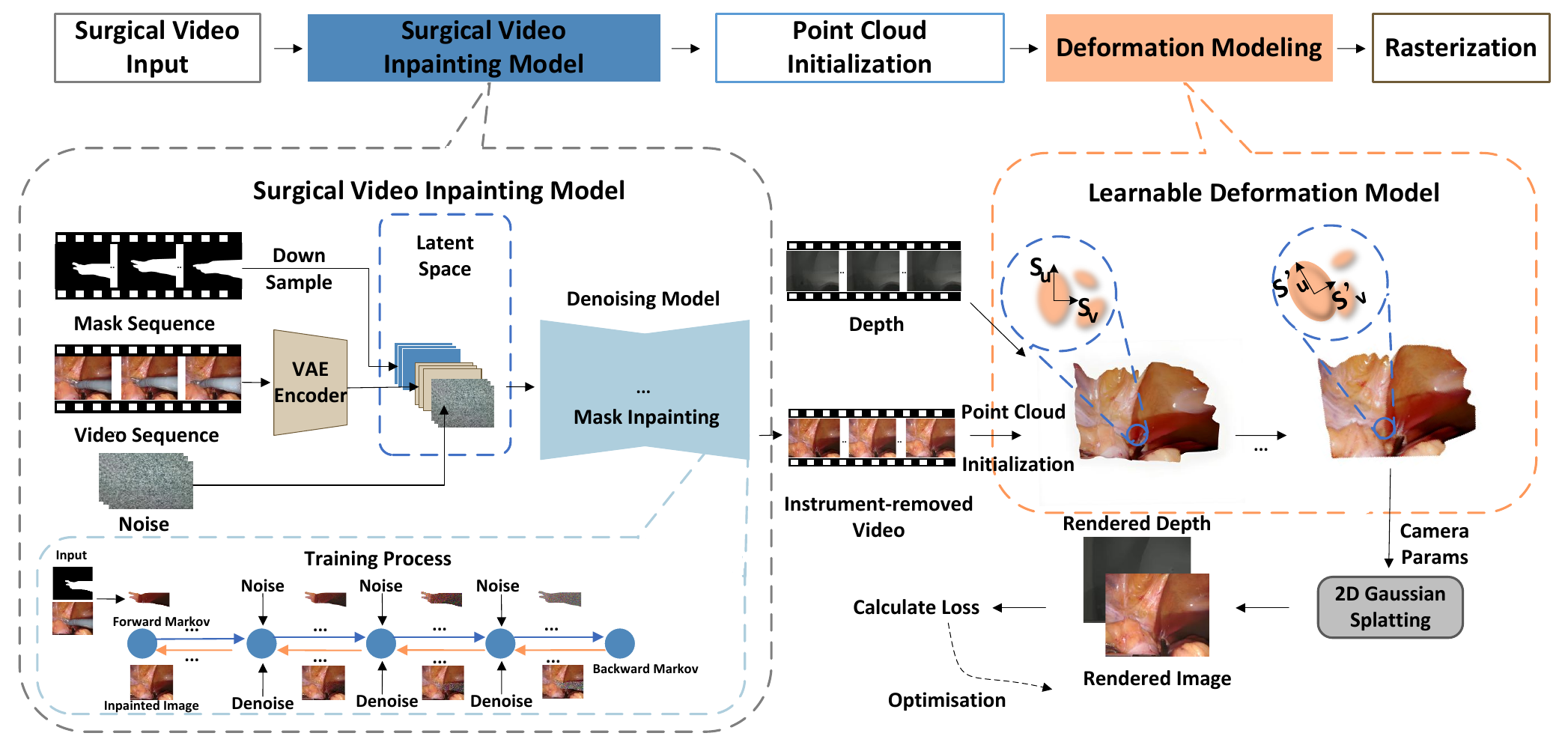}

\caption{Our reliable surgical scene reconstruction framework, Diff2DGS, consists of Surgical Instrument Inpainting, Point Cloud Initialization, Deformation Modeling, and 2D Gaussian Splatting.}
\label{fig:framework}
\vspace{-0.3cm}
\end{figure*}

\subsection{Occlusion and evaluation gaps}
Surgical tool occlusions are a major obstacle to reliable reconstruction because the underlying tissue appearance and geometry are not directly observed. Existing methods commonly mask out instrument regions during optimization, which avoids penalizing occluded pixels but leaves the hidden tissue geometry under-constrained. In addition, evaluation is often performed on EndoNeRF and StereoMIS, which lack ground-truth 3D geometry and therefore rely mainly on image-space metrics such as PSNR and SSIM. As illustrated in Fig.~\ref{fig:single_column}, high image quality from the camera viewpoint does not necessarily imply geometrically accurate reconstruction from novel viewpoints. Although no public surgical video dataset provides both severe tissue deformation and complete 3D ground truth, SCARED \cite{scared} provides stereo robotic surgery videos with structured-light 3D anatomy ground truth, making it useful for additional geometry-oriented evaluation.

Tool removal also requires temporally consistent restoration of occluded tissue regions. Recent video restoration methods exploit diffusion models \cite{diff1,diff2}, but diffusion-based inpainting can introduce hallucination artifacts if temporal consistency is not sufficiently constrained. Methods such as FloED \cite{floed}, FFF-VDI \cite{fff_vdi}, and DiffuEraser \cite{diffueraser} therefore incorporate temporal priors to improve video consistency and reduce hallucinations.

\subsection{Our approach}
Together, these challenges motivate an efficient deformable reconstruction framework that is robust to surgical tool occlusions and explicitly evaluated for geometric reliability. 
This paper proposes Diff2DGS, a two-stage framework for deformable intraoperative 3D reconstruction. First, we enhance diffusion-based mask inpainting by integrating temporal video priors to restore tissue occluded by surgical instruments. Second, we extend 2D Gaussian Splatting (2DGS) \cite{2dgs0} to surgical scenes using a Learnable Deformation Model (LDM) that accounts for dynamic tissue deformation and anatomical geometry. We further optimize reconstruction geometry using an adaptive depth loss. We evaluate Diff2DGS on three publicly available da Vinci robotic surgery datasets: EndoNeRF \cite{surgey_nerf0}, StereoMIS \cite{stereomis}, and SCARED~\cite{scared}. Overall, our main contributions are as follows:

\begin{itemize}
    \item We present Diff2DGS, a two-stage framework that restores tissue regions occluded by surgical instruments before deformable 3D reconstruction, thereby reducing artifacts in occluded regions.

    \item We extend the 2D Gaussian representation to deformable surgical tissue reconstruction and introduce an LDM to efficiently model tissue motion.

    \item We propose an adaptive depth loss that dynamically balances RGB reconstruction and depth supervision to improve geometric accuracy.

    \item We validate Diff2DGS through quantitative and qualitative evaluations on EndoNeRF, StereoMIS, and SCARED, achieving 38.02 dB PSNR on EndoNeRF and 34.40 dB PSNR on StereoMIS.
\end{itemize}

\section{Method}

The overall framework of Diff2DGS is depicted in Fig.~\ref{fig:framework}. First, we use a diffusion model to remove surgical instruments and inpaint them with the tissue underneath. This process utilizes surgical instrument segmentation across frames and generates high-quality tissue images with spatiotemporal consistency using Stable Diffusion and temporal attention (Section~2.3). Inpainted images are then combined with stereo depth information to initialize a Gaussian point cloud, and tissue deformation is modeled by the LDM (Section~2.2). Subsequently, a 2DGS model is applied to generate a color image and depth map from the perspective of the given camera, which is further refined through optimisation.

\subsection{Preliminaries of Gaussian Splatting}

3D Gaussian Splatting (3DGS) was originally proposed to model static scenes \cite{3dgs0} as a set of 3D Gaussian intensity distributions that can be projected to arbitrary 2D viewpoints. Each Gaussian is characterized by its 3D position $\boldsymbol{\mu}\in\mathbb{R}^3$ and $3\times3$ covariance matrix $\boldsymbol\Sigma$ encoding its scale ($\mathbf S$) and orientation ($\mathbf R$) in world coordinates. For any 3D point $\mathbf{x}=(x,y,z)^\top\in\mathbb{R}^3$ in world coordinates, the impact of a 3D Gaussian on $\mathbf{x}$ is defined by the Gaussian distribution:
\begin{equation}
{
G^{3D}(\mathbf{x})=\exp \left(-\frac{1}{2}(\mathbf{x}-\boldsymbol{\mu})^\top (\boldsymbol{\Sigma}^{3D})^{-1}(\mathbf{x}-\boldsymbol{\mu})\right)
}
\end{equation}

\begin{equation}
 {
\boldsymbol{\Sigma}^{3D}=\mathbf{R}\mathbf{S}\mathbf{S}^\top\mathbf{R}^\top
}
\end{equation}

Given the world-to-camera transformation matrix $\mathbf{W}$, in the rendering process, the covariance matrix can be projected into image space as:

\begin{equation}
 {
\boldsymbol{\Sigma}'^{3D} = \mathbf{J}\mathbf{W}\boldsymbol{\Sigma}^{3D} \mathbf{W}^{\top} \mathbf{J}^{\top}
}
\label{eq:w2c}
\end{equation}
{\noindent where $\mathbf{J}$ is the Jacobian of the affine approximation of the projective transformation. However, 3D Gaussians are often suboptimal as a representation of surface textures and edges. To mitigate this, a 2D Gaussian \cite{2dgs0} representation proposes instead to model a scene in terms of planar Gaussians embedded in 3D space. Each 2D Gaussian is represented by a central point $\mathbf{p}_c$, two orthogonal tangent vectors $\mathbf{t}_u$ and $\mathbf{t}_v$}, and two scalar scale parameters $s_u$ and $s_v$ along the tangent directions. Any 3D point $\mathbf{P}$ in world coordinates can be parametrised in terms of $u,v$ 2D coordinates in the tangent plane:

\begin{table*}[h]
\caption{Comparison between Diff2DGS and Existing Methods on EndoNeRF.}
\centering
\resizebox{\textwidth}{!}{
\begin{tabular}{c|c|c|c|c|c|c|c|c|c}
\hline
\multirow{2}{*}{Models} & \multicolumn{4}{c|}{EndoNeRF-Cutting} & \multicolumn{4}{c|}{EndoNeRF-Pulling} & \multirow{2}{*}{FPS}\\
\cline{2-9}
 & PSNR $\uparrow$ & SSIM (\%) $\uparrow$ & LPIPS $\downarrow$ & RMSE $\downarrow$ & PSNR $\uparrow$ & SSIM (\%) $\uparrow$ & LPIPS $\downarrow$ & RMSE $\downarrow$ & \\
\hline

EndoNeRF \cite{surgey_nerf0} & 35.84 & 94.25 & 0.085 & 4.299 & 35.43 & 93.89 & 0.077 & 4.746 & 0.03 \\
LerPlane \cite{dynamic1} & 35.21 & 92.96 & 0.095 & 5.782 & 34.87 & 91.94 & 0.118 & 6.245 & 0.93 \\
EndoGaussian \cite{endogaussian} & 37.19 & 95.29 & 0.052 & 4.032 & 35.98 & 94.23 & \underline{0.047} & 5.341 & 83.13 \\
Deform3DGS \cite{deform3dgs} & 37.33 & 95.89 & 0.075 & 4.234 & \underline{37.70} & \underline{95.57} & 0.069 & 5.727 & \underline{228.58} \\
SurgicalGS \cite{surgicalgs} & \underline{37.93} & \underline{96.19} & \underline{0.045} & \textbf{1.779} & 37.13 & 95.43 & 0.062 & \textbf{2.193} & 154.39 \\
Diff2DGS (Ours) & \textbf{38.23} & \textbf{96.76} & \textbf{0.033} & \underline{2.443} & \textbf{37.81} & \textbf{96.12} & \textbf{0.042} & \underline{2.574} & \textbf{232.29} \\
\hline
\end{tabular}}
\label{endonerfcompare}
\end{table*}

\begin{table*}[h]
\caption{Comparison between Diff2DGS and Existing Methods on StereoMIS.}
\centering
\resizebox{\textwidth}{!}{
\begin{tabular}{c|c|c|c|c|c|c|c|c|c}
\hline
\multirow{2}{*}{Models} & \multicolumn{4}{c|}{StereoMIS-P2\underline{ }7} & \multicolumn{4}{c|}{StereoMIS-P3} & \multirow{2}{*}{FPS}\\
\cline{2-9}
 & PSNR $\uparrow$ & SSIM (\%) $\uparrow$ & LPIPS $\downarrow$ & RMSE $\downarrow$ & PSNR $\uparrow$ & SSIM (\%) $\uparrow$ & LPIPS $\downarrow$ & RMSE $\downarrow$ & \\
\hline

\text {EndoNeRF \cite{surgey_nerf0}}   & 28.87   &  80.23  & 0.259 & 4.354 & 28.33  & 80.09 & 0.287 & 4.165 & 0.06  \\
 \text { {LerPlane}\cite{dynamic1} }    &   29.53     & 78.36    & 0.205 & 8.356 &29.41 &78.14 &0.213 & 7.983 & 0.97\\
 \text { EndoGaussian \cite{endogaussian} } & 31.11      & 87.29    & \underline{0.127} & 9.793 &31.03 &87.13 &0.136  & 8.932 & 90.31\\
 \text { Deform3DGS\cite{deform3dgs} } & \underline{31.83} &87.75 & 0.129 & 10.036 &30.91  & 87.73 & \underline{0.131} & 8.482 &  \textbf {214.38} \\
 \text { SurgicalGS\cite{surgicalgs} } &31.63 & \underline{88.26}  & 0.152 & \textbf {2.747} & \underline{31.49}  &\underline{87.75} & 0.165 & \textbf {2.536} &154.82 \\
 \text { Diff2DGS (Ours) }     & \textbf {34.72}  & \textbf {91.17} & \textbf {0.113} & \underline{3.874} & \textbf {34.08} &\textbf {90.27}  & \textbf {0.131} & \underline{3.717} & \underline{212.32}\\
\hline
\end{tabular}}
\label{stereomiscompare}

\end{table*}

\begin{equation}
\mathbf{P}(u, v) = \mathbf{p}_c + s_u \mathbf{t}_u u + s_v \mathbf{t}_v v
= \mathbf{H}(u, v, 0, 1)^{\top}
\end{equation}

\begin{equation}
 {
\mathbf{H} =
\begin{bmatrix}
s_u \mathbf{t}_u & s_v \mathbf{t}_v & \mathbf{0} & \mathbf{p}_c \\
0 & 0 & 0 & 1
\end{bmatrix}
=
\begin{bmatrix}
\mathbf{R}\mathbf{S} & \mathbf{p}_c \\
0 & 1
\end{bmatrix}}
\label{eq:susvtutv}
\end{equation}
\noindent where $\mathbf{H}$ is a homogeneous transformation matrix representing the geometry of the 2D Gaussian. $\mathbf{R}$ is a $3 \times 3$ rotation matrix $[\mathbf{t}_u,\mathbf{t}_v,\mathbf{t}_u\times \mathbf{t}_v]$, and $\mathbf{S}$ is a $3 \times 3$ diagonal matrix whose last entry is zero.

The 2D Gaussian $G^{2D}$ can be represented by a $2 \times 2$ covariance matrix $\boldsymbol{\Sigma}'^{2D}$ in $uv$ space, obtained by removing the third row and column of $\boldsymbol{\Sigma}'^{3D}$ in \eqref{eq:w2c}. 
For an image-plane coordinate $\mathbf{q}$, the corresponding camera ray can be written as $\mathbf{r}(\mathbf{q},\tau)=\mathbf{o}+\tau\mathbf{d}(\mathbf{q})$, where $\mathbf{o}$ is the camera center and $\mathbf{d}(\mathbf{q})$ is the ray direction. For the $i$-th 2D Gaussian, this ray intersects its tangent plane at a 3D point that can be expressed by the local coordinate $\mathbf{u}_i(\mathbf{q})=(u_i,v_i)^\top$ through Eq.~\eqref{eq:susvtutv}. The 2D Gaussian value is therefore evaluated in this local tangent-plane coordinate:
\begin{equation}
 {
  G_i^{2D}(\mathbf{u}_i(\mathbf{q}))
  =
  \exp\!\left(
    -\tfrac{1}{2}\,
    \mathbf{u}_i(\mathbf{q})^\top
    \left(\boldsymbol{\Sigma}'^{2D}\right)^{-1}
    \mathbf{u}_i(\mathbf{q})
  \right).
  \label{eq:2d-gaussian-standard}
}
\end{equation}

The rasterization process in 2DGS is similar to that in 3DGS, both employing volumetric alpha blending to integrate alpha-weighted appearance from front to back. For an image-plane coordinate $\mathbf{q}$, the rendered color is:
\begin{equation}
 {
C(\mathbf{q})=\sum_{i=1}^{N} G_i^{2D}(\mathbf{u}_i(\mathbf{q})) \alpha_i c_i 
\prod_{j=1}^{i-1}\left(1-G_j^{2D}(\mathbf{u}_j(\mathbf{q})) \alpha_j\right)
}
\end{equation}
  
\noindent where $\alpha_i$ represents the opacity and $c_i$ denotes the color of each Gaussian primitive.



\subsection{Learnable Deformation Model for 2D Gaussian Splatting(LDM)}
The original 2DGS is limited to reconstructing static scenes. Therefore, we propose a learnable deformation model (LDM) that follows a similar strategy to Deform3DGS\cite{deform3dgs}, but utilises a 2D Gaussian representation for better reconstruction of tissue surfaces. Specifically, we employ Gaussian functions with learnable centers $\theta$ and variances $\sigma$ for estimating tissue deformation.
\begin{equation}
\tilde{G'}(t ; \theta, \sigma)=\exp \left(-\frac{1}{2 \sigma^2}(t-\theta)^2\right)
\end{equation}
Here, $t$ denotes time. For each Gaussian $g_i$ in the point cloud, the position $\mathbf{p}_c$ and the rotation $R$ are directly related to tissue deformation. The scale $S$ varies as tissues experience elastic deformations during instrument intervention. Therefore, we have established a set of learned parameters {$ \Theta^S , \Theta^R , \Theta^{\mathbf p_c}$} to describe deformation through scale, rotation, and position. As described in Eq.~\eqref{eq:susvtutv}, we map the matrices $R$ and $S$ to the 2D parameters $s_u$, $\mathbf{t}_u$, $s_v$, and $\mathbf{t}_v$. Consequently, $\Theta^S$ and $\Theta^R$ can be further decomposed into $\Theta^{s_u}$, $\Theta^{s_v}$, $\Theta^{\mathbf{t}_u}$, and $\Theta^{\mathbf{t}_v}$. Taking the scale$S$ change in $u$ direction  as an example, the $\Theta^{s_u}$ can be defined as the deformation of $S$ in $u$ direction, the new scale in $u$ direction $s_u'$ is:
\begin{equation}
{
s_u'= s_u + \Theta^{s_u} = s_u + \sum_{j=1}^B \omega_j^{s_u} \tilde{G'}\left(t ; \theta_j^{s_u}, \sigma_j^{s_u}\right)
}
\end{equation}
where $B$ is the number of temporal Gaussian basis functions, and $\omega_j^{s_u}$ denotes the learnable coefficient of the $j$-th basis function for the scale deformation along the $u$ direction.
By integrating the deformation variables, the LDM ensures smooth transitions and continuity in deformation across adjacent time points, maintaining a coherent and consistent temporal progression.


\subsection{Diffusion-based Inpainting of Surgical Instruments}

Tissue 3D reconstruction is made challenging by the presence of surgical instruments that partially occlude the view. 
Previous surgical reconstruction methods commonly use instrument segmentation masks to exclude tool regions from the photometric or reconstruction loss during NeRF- or Gaussian-Splatting-based optimization \cite{surgey_nerf0,deform3dgs,endogaussian}.
We instead propose to use a diffusion model \cite{ddim} as a pre-processing step to remove surgical instrument occlusions and inpaint them with the tissue appearance underneath. We then perform 3D reconstruction without masking out any region.

The video inpainting model categorizes pixels into two types: known pixels (non-occluded tissue) that are propagated using temporal information from preceding frames, and unknown pixels (occluded tissue) that are generated by the diffusion model while preserving structural integrity. To further enhance consistency in long video sequences, a temporal attention mechanism is utilised to extend the model’s temporal receptive field. The temporal attention mechanism $TA_t$ can be defined as:

\begin{equation}
TA_t = \mathrm{Softmax}\!\left(\mathbf{Q}_t\mathbf{K}_t^\top/\sqrt{d_k} + \mathbf{M}_t\right)\mathbf{V}_t.
\end{equation}
Here $\mathbf{Q}_t,\mathbf{K}_t,\mathbf{V}_t$ are linearly projected features from latent space, $d_k$ denotes the head dimension, and $\mathbf{M}_t$ is a causal mask\cite{causalstory} with $(\mathbf{M}_t)_{ij}=0$ if $i\ge j$ and $(\mathbf{M}_t)_{ij}=-\infty$ otherwise.

 {
We optimize the diffusion-based inpainting module using a mask-weighted L2 loss in the latent space.
Given a clean latent $\mathbf{z}_0$ and timestep $t$, we construct a noisy latent $\mathbf{z}_t$ by the forward diffusion process, i.e.,
$\mathbf{z}_t = \sqrt{\bar{\alpha}_t}\mathbf{z}_0 + \sqrt{1-\bar{\alpha}_t}\boldsymbol{\epsilon}$ with $\boldsymbol{\epsilon}\sim\mathcal{N}(\mathbf{0},\mathbf{I})$.
The denoising network $\hat{\boldsymbol{\epsilon}}_{\theta}$ predicts the injected noise conditioned on the masked inputs and temporal priors, denoted by $\mathbf{c}$.
Let $\mathbf{m}$ be a binary inpainting mask downsampled to the latent resolution (with $\mathbf{m}=1$ indicating the occluded/inpainted region). The training loss is:
\begin{equation}
\mathcal{L}_{\text{inpaint}}
=
\mathbb{E}_{\mathbf{z}_0,t,\boldsymbol{\epsilon}}
\left[
\left\|
\mathbf{m}\odot
\Big(
\boldsymbol{\epsilon}-\hat{\boldsymbol{\epsilon}}_{\theta}(\mathbf{z}_t,t,\mathbf{c})
\Big)
\right\|_2^2
\right],
\end{equation}
where $\odot$ denotes the Hadamard product. This objective emphasizes accurate restoration inside the masked region while maintaining global structural coherence.
}

 {
For inference, we adopt the DDIM~\cite{ddim} sampling strategy. We first estimate the clean latent $\hat{\mathbf{z}}_0$ at each step:
\begin{equation}
\hat{\mathbf{z}}_0 = \big(\mathbf{z}_t - \sqrt{1 - \bar{\alpha}_t} \hat{\boldsymbol{\epsilon}}_{\theta}(\mathbf{z}_t, t, \mathbf{c})\big) / \sqrt{\bar{\alpha}_t}.
\end{equation}
Then, the latent at the previous timestep $\mathbf{z}_{t-1}$ is computed as: 
\begin{equation}
\mathbf{z}_{t-1} = \sqrt{\bar{\alpha}_{t-1}} \hat{\mathbf{z}}_0 + \sqrt{1 - \bar{\alpha}_{t-1}} \hat{\boldsymbol{\epsilon}}_{\theta}(\mathbf{z}_t, t, \mathbf{c}).
\end{equation}}

\begin{figure*}
\centering
\includegraphics[width=0.85\textwidth]{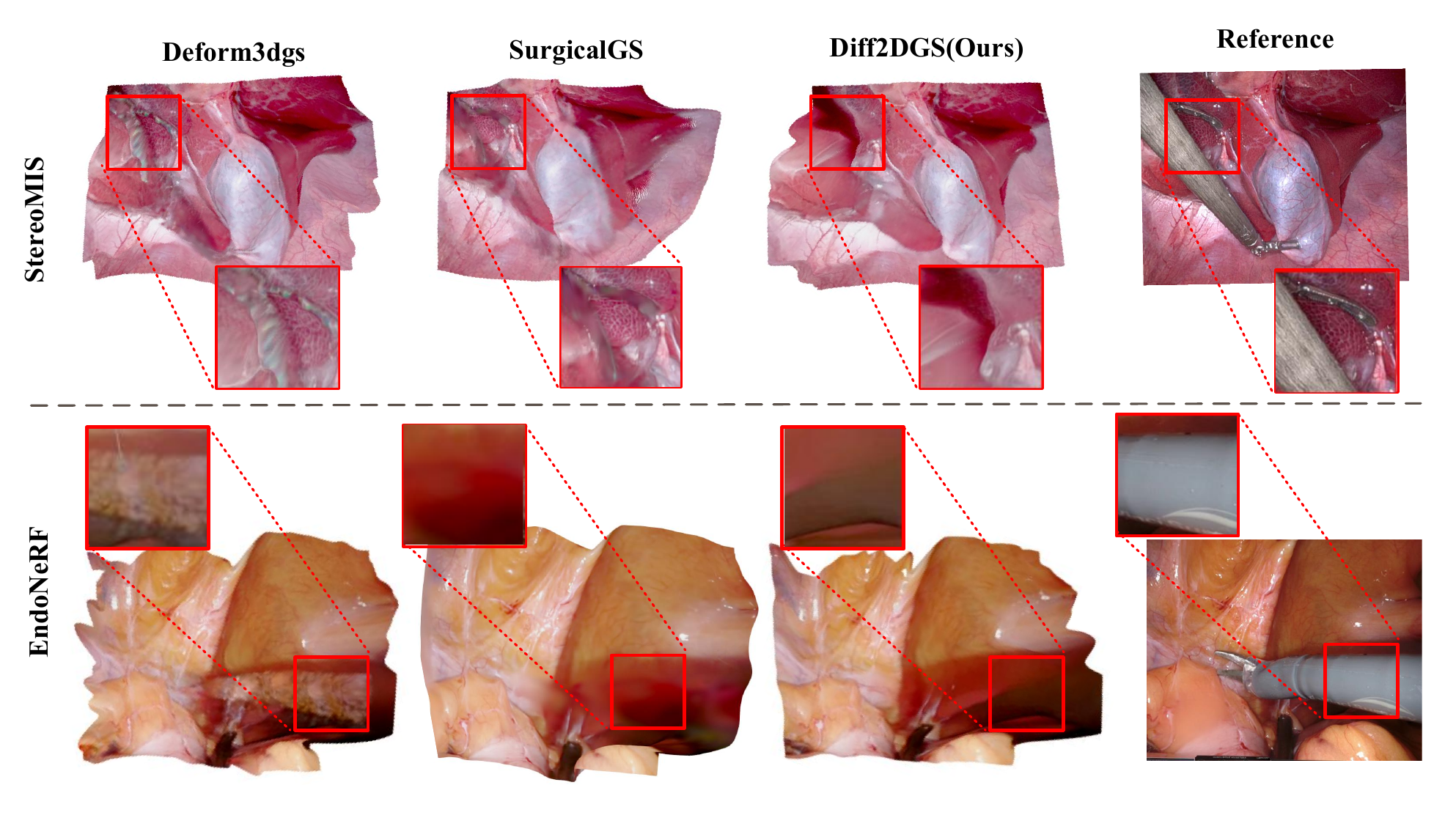}
\vspace{-0.2cm}
\caption{Qualitative comparison of 3D reconstruction results on the
StereoMIS and EndoNeRF datasets.} \label{fig3}
\vspace{-0.3cm}
\end{figure*}

\subsection{Adaptive Depth Loss Weight}

As shown in Fig.~\ref{fig:single_column}, Gaussian Splatting often produces reconstructions with sharp depth changes that do not correspond to an accurate representation of the tissue shape. To enhance the model's attention to depth accuracy, we adopt an adaptive depth loss weighting strategy that dynamically adjusts the weight of the depth loss throughout the training process.

Specifically, in 3D reconstruction tasks, we aim to simultaneously minimize the RGB reconstruction error $L_{\mathrm{rgb}}$ and the depth reconstruction error $L_{\mathrm{depth}}$:
\begin{equation}
\mathcal{L}=L_{\mathrm{rgb}}+\lambda_{\mathrm{depth}} L_{\mathrm{depth}} .
\end{equation}
Here, $L_{\mathrm{rgb}}=\|\hat{I}-I^*\|_1$ denotes the L1 reconstruction loss between the rendered image $\hat{I}$ and reference image $I^*$. The depth term is defined as an inverse-depth L1 loss:
\begin{equation}
L_{\mathrm{depth}}=\frac{1}{|\Omega_d|}\sum_{\mathbf{q}\in\Omega_d}
\left|\hat{D}(\mathbf{q})^{-1}-D^{*}(\mathbf{q})^{-1}\right|,
\end{equation}
where $\mathbf{q}$ denotes an image-plane coordinate, $\hat{D}$ and $D^{*}$ are the rendered and reference depth maps, respectively, and $\Omega_d$ contains pixels with valid non-zero depth.

In a conventional fixed-weight formulation, $\lambda_{\mathrm{depth}}$ is manually set. However, using fixed weights makes it difficult to balance the dynamic range and convergence speed of both losses. In the early stages of training, the RGB loss often dominates, whereas in the later stages, the depth loss may become more significant. Rather than eliminating hyperparameters, our strategy dynamically updates the effective depth weight according to the current ratio between $L_{\mathrm{rgb}}$ and $L_{\mathrm{depth}}$ during optimization.

To prevent the depth term from becoming overly dominant in the later stages of training, we first define a base weight that decays linearly with the number of iterations:
\begin{equation}
w_{\text{base}}(t)=w_{\text{init}}\left(1-\alpha t/T\right), \quad 0 \leq t \leq T,
\end{equation}
where $w_{\text{init}}$ is the initial weight, $T$ is the total number of iterations, and $\alpha \in[0,1]$ controls the attenuation amplitude.

We dynamically adjust the depth loss weight based on the loss ratio:
\begin{equation}
r(t)=L_{\mathrm{rgb}}(t)/(L_{\mathrm{depth}}(t)+\varepsilon), \quad \varepsilon=10^{-8}.
\end{equation}
When $r(t)>1$, it indicates that the RGB loss is greater than the depth loss, and the depth weight should be increased; otherwise, it should be decreased. To achieve smooth adjustment, we introduce the hyperbolic tangent function:
\begin{equation}
g(r(t))=1+\beta \tanh (r(t)-1), \quad \beta \in(0,1].
\end{equation}
The final weight is:
\begin{equation}
\lambda_{\text{depth}}(t)=\operatorname{clip}\left(w_{\text{base}}(t) g(r(t)), w_{\min}, w_{\max}\right).
\end{equation}
Here, $\operatorname{clip}(x, a, b)=\min(\max(x, a), b)$. The weights are constrained to $[w_{\min}, w_{\max}]$, where $w_{\min}$ and $w_{\max}$ are fixed lower and upper bounds. In our experiments, we use one global setting for all sequences: $w_{\mathrm{init}}=10$, $\alpha=0.8$, $\beta=0.25$, $w_{\min}=1$, and $w_{\max}=10$.

\section{Experiments}
\subsection{Experiment Setting}

\textbf{Datasets and Evaluation:} We evaluate the proposed method and compare it with existing works on EndoNeRF~\cite{surgey_nerf0}, StereoMIS~\cite{stereomis} and SCARED~\cite{scared} datasets. 

The EndoNeRF is a collection of stereo endoscopic videos, including 6 clips extracted from Da Vinci robotic prostatectomy data. The StereoMIS is a stereo endoscopic video dataset captured from in-vivo porcine subjects, containing diverse anatomical structures and challenging scenes with large tissue deformations.  The masks of EndoNeRF and StereoMIS are both provided in the datasets. 
For both of these datasets, we evaluate methods using image similarity metrics (PSNR, SSIM, LPIPS) in line with previous literature using these datasets~\cite{deform3dgs,endogaussian}. However, our results suggest that image similarity is not sufficient to fully evaluate 3D reconstruction accuracy and should be complemented with geometry-based metrics. Given that neither of these datasets has 3D shape groundtruth, we provide depth difference (RMSE) between methods' estimates and stereo reconstructions (RAFT~\cite{raft}). Additionally, we evaluate our method on the SCARED dataset, which has depth groundtruth obtained with a structured light sensor. It contains images of five porcine cadaver abdominal anatomies acquired with a DaVinci endoscope. Unlike EndoGaussian~\cite{endogaussian}, which evaluates only the regions overlapping with a reference view by masking newly visible areas introduced by camera motion, we report metrics on the entire reconstructed scene. For a fair comparison, all methods including EndoGaussian, are re-evaluated under this unified protocol.

\textbf{Experiment Details:} 
The inpainting module is built upon Stable Diffusion v1.5 and comprises approximately 2.47B parameters in total. We use Phased Consistency Model (PCM\cite{pcm}) to accelerate the diffusion inference stage, which only requires two denoising steps per frame to perform mask inpainting. The model is fine-tuned on surgical sequences of SCARED dataset where supervision is provided by a mask-weighted $L_2$ loss with the learning rate of 1e-5. 
For the Gaussian Splatting module, we optimize one scene-specific model for each video sequence. The video duration of each sequence is normalized to $[0,1]$. Following prior surgical scene reconstruction protocols, every eighth frame is held out for testing, while the remaining frames are used for optimisation. The training runs for 6000 iterations with an initial learning rate of $3.2 \times 10^{-3}$. To stabilize training, we freeze the number of Gaussian points during the initial 600 iterations.
All experiments are implemented using the PyTorch framework\cite{pytorch} and conducted on a single NVIDIA Tesla V100 GPU.


\begin{table*}[t]
\centering
\scriptsize
\caption{Quantitative comparison of different methods on SCARED dataset.}
\label{scaredcompare}
\begin{tabular}{@{}lccccccccc@{}}
\toprule
\multirow{2}{*}{\textbf{Method}} & 
\multicolumn{3}{c}{\textbf{Full Image}} & 
\multicolumn{3}{c}{\textbf{Masked Region}} & 
\multirow{2}{*}{\textbf{SSIM}$(\%)$$\uparrow$} &
\multirow{2}{*}{\textbf{FID}$\downarrow$} &
\multirow{2}{*}{\textbf{TCS}$\downarrow$} \\ 
\cmidrule(lr){2-4} \cmidrule(lr){5-7}
 & PSNR$\uparrow$ & LPIPS$\downarrow$ & RMSE$\downarrow$ 
 & PSNR$\uparrow$ & LPIPS$\downarrow$ & RMSE$\downarrow$ 
 & & & \\
\midrule
Inpaint\_result & \textbf{31.84} & \textbf{0.100} & \textbf{6.68} & \underline{27.01} & \textbf{0.017} & \underline{12.05} & \textbf{90.24} & \textbf{29.92} & \textbf{0.003} \\
Deform3DGS      & 25.34 & 0.342 & 13.83 & 19.52 & 0.037 & 27.19 & 72.61 & 74.16 & 0.027 \\
EndoGaussian      & 25.96 & 0.398 & 12.89 & 23.56 & 0.032 & 17.61 & 68.39 & 81.95 & 0.029 \\
SurgicalGS      & 24.92 & 0.434 & 14.51 & 21.63 & 0.040 & 21.57 & 65.53 & 108.89 & 0.032 \\
Diff2DGS        & \underline{27.77} & \underline{0.314} & \underline{10.53} & \textbf{30.53} & \underline{0.022} & \textbf{8.21} & \underline{76.14} & \underline{61.02} & \underline{0.027} \\
\bottomrule
\end{tabular}
\vspace{-0.3cm}
\end{table*}

\subsection{Comparison with State-of-the-art Methods}
To ensure a fair and consistent comparison, we evaluate the proposed Diff2DGS framework alongside existing state-of-the-art methods under the same experimental environment and hardware configurations.
As shown in Table~\ref{endonerfcompare} and Table~\ref{stereomiscompare}, our framework outperforms Deform3DGS \cite{deform3dgs} on both EndoNeRF and StereoMIS datasets, demonstrating the value of a 2D Gaussian parametrisation. 
SurgicalGS achieves lower RMSE because it explicitly uses stereo-depth priors, including multi-frame depth fusion and depth regularization. Since RMSE on EndoNeRF and StereoMIS is measured against RAFT-based stereo depth rather than ground-truth geometry, methods that directly incorporate stereo-depth estimates are naturally favored under this metric. Therefore, the lower RMSE of SurgicalGS should be interpreted as stronger agreement with the stereo-depth reference, rather than necessarily more accurate ground-truth geometry. In contrast, Diff2DGS achieves better image fidelity and faster rendering, while obtaining the second-best RMSE among the compared methods. As a Gaussian splatting method, our framework's rendering speed is hundreds of times faster than NeRF-based methods such as EndoNeRF and LerPlane. When compared to other Gaussian splatting methods, Diff2DGS achieves rendering speeds comparable to Deform3DGS while improving reconstruction results in occluded regions. As a result, our framework provides a favourable balance between reconstruction quality and rendering speed. 

Fig.~\ref{fig3} shows rendered surgical scenes for visual evaluation comparing our method with Deform3DGS and SurgicalGS. We observe that both Deform3DGS and SurgicalGS exhibit significant artifacts in occluded regions. In contrast, our method achieves superior image similarity metrics primarily due to more accurate reconstruction of instrument-occluded areas, 
demonstrating the effectiveness of diffusion-based inpainting.

To validate performance in occluded regions, we overlay synthetic surgical instrument masks on SCARED and use the original images as ground truth.

As shown in Table~\ref{scaredcompare} and Fig.~\ref{fig:inpaint-scared}, Diff2DGS achieves 30.53 dB PSNR and 8.21 mm RMSE in masked regions, outperforming Deform3DGS, EndoGaussian, SurgicalGS, and the inpainting baseline.
This protocol enables quantitative evaluation with known tissue ground truth, but may not fully capture real surgical effects such as specular highlights, motion blur, soft shadows, or imperfect mask boundaries; these gaps can be mitigated by target-domain fine-tuning with real surgical videos and masks.

To assess temporal stability, we define the Temporal Consistency Score as $\mathrm{TCS}=|\mathrm{TC}_{\mathrm{result}}-\mathrm{TC}_{\mathrm{GT}}|$, where TC is the average RMSE between consecutive frame pairs; lower TCS indicates better temporal consistency. FID is reported as a unitless distributional image-quality metric, where lower is better. As shown in Table~\ref{scaredcompare}, the inpainting module achieves the best temporal consistency and FID score, while Diff2DGS obtains the second-best result and remains superior to the other reconstruction methods.

\begin{figure}[htbp]
  \centering
  \includegraphics[width=0.9\columnwidth]{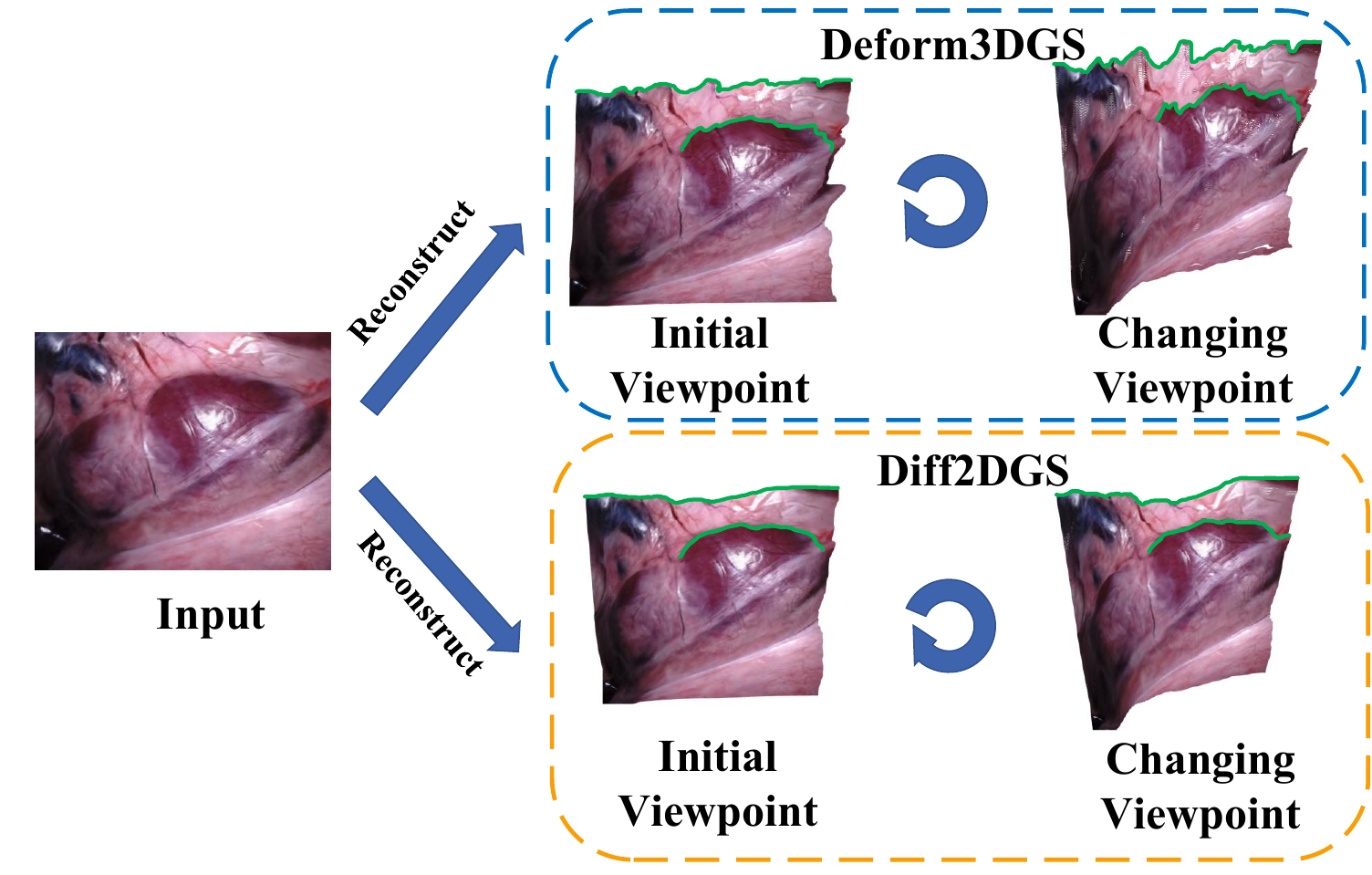}
  \caption{Depth quality visualization on the SCARED dataset. Green contours highlight visible depth inconsistency near tissue boundaries.}
  \vspace{-0.4cm}
  
  \label{fig:scaredcompare}
\end{figure}
\begin{figure*}
\centering
\includegraphics[width=0.85\textwidth]{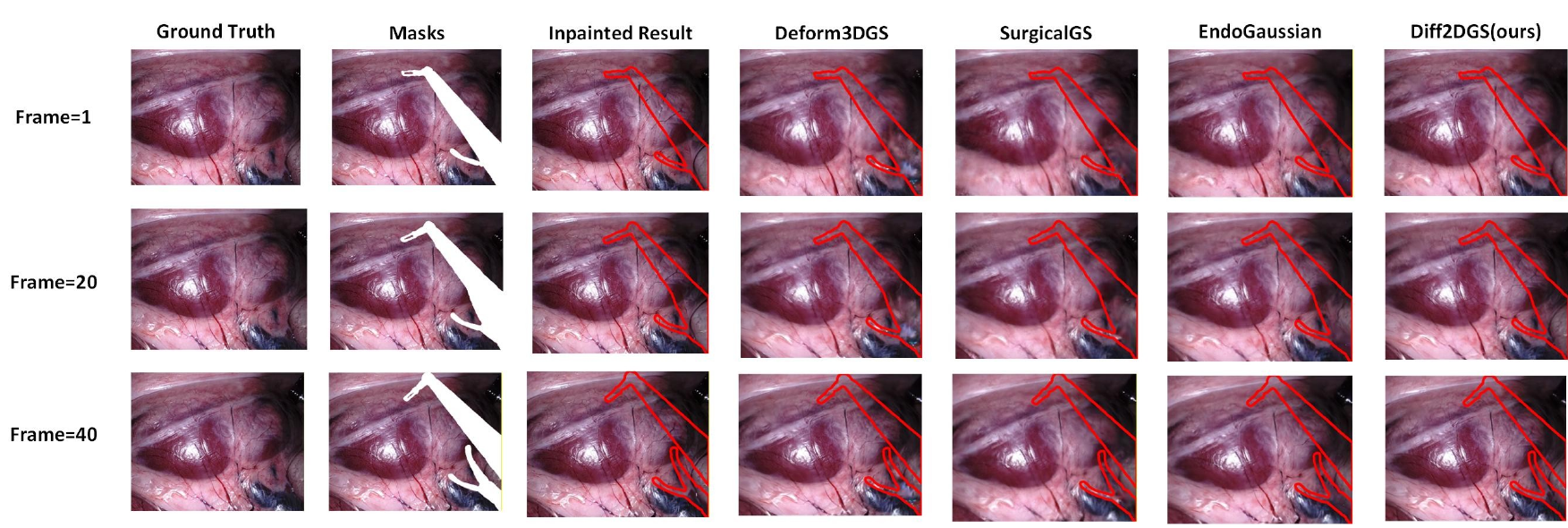}
\vspace{-0.2cm}
\caption{Qualitative comparison of reconstructed instrument-occluded
regions on the SCARED dataset.} 
\vspace{-0.3cm}
\label{fig:inpaint-scared}
\end{figure*}

Fig.~\ref{fig:scaredcompare} compares the reconstructed surfaces rendered from the initial viewpoint and a novel viewpoint. The added camera markers and 3D coordinate axes indicate the viewpoint change, while the green contours highlight regions with visible geometric artifacts. When rendered from the novel viewpoint, Deform3DGS exhibits sharper and more irregular surface discontinuities near tissue boundaries. These discontinuities lead to inconsistent depth changes, indicating poorer depth quality and less reliable geometry. In contrast, Diff2DGS preserves smoother surface structure in the same regions, indicating improved geometric consistency.

\subsection{Ablation experiment}
To verify the effectiveness of each module in Diff2DGS, we designed several variant models and conducted experiments on both the StereoMIS and EndoNeRF datasets. The results are shown in Table~\ref{ablationstudy}.

\begin{table}[t]
\centering
\caption{Ablation study of Diff2DGS.}
\resizebox{\columnwidth}{!}{
\begin{tabular}{>{\raggedright\arraybackslash}p{2.7cm}|
                >{\centering\arraybackslash}p{0.5cm}
                >{\centering\arraybackslash}p{0.9cm}
                >{\centering\arraybackslash}p{0.7cm}|
                >{\centering\arraybackslash}p{0.5cm}
                >{\centering\arraybackslash}p{0.9cm}
                >{\centering\arraybackslash}p{0.7cm}}
\hline
\multirow{2}{*}{Models}
& \multicolumn{3}{c|}{StereoMIS}
& \multicolumn{3}{c}{EndoNeRF} \\
& PSNR$\uparrow$ & SSIM(\%)$\uparrow$ & RMSE$\downarrow$
& PSNR$\uparrow$ & SSIM(\%)$\uparrow$ & RMSE$\downarrow$ \\
\hline
Diffusion+Deform3DGS
& 34.32 & 90.13 & 8.13
& 37.63 & 95.76 & 4.78 \\

w/o deformation
& 32.12 & 86.87 & 6.30
& 33.94 & 89.85 & 5.74 \\

w/o inpainting
& 31.95 & 86.11 & 4.39
& 37.57 & 95.73 & 2.65 \\

 w/o $L_{\mathrm{depth}}$ ($\lambda=0$)
&  33.76 &  90.14 &  9.80
&  36.95 &  95.41 &  7.37 \\

w/o adaptive ($\lambda=1$)
&  33.99 &  90.21 &  8.58
&  37.86 &  95.72 &  6.74 \\

Diff2DGS (Ours)
& \textbf{34.40} & \textbf{90.72} & \textbf{3.79}
& \textbf{38.02} & \textbf{96.44} & \textbf{2.51} \\
\hline
\end{tabular}}
\label{ablationstudy}
\vspace{-0.3cm}
\end{table}

To validate the effectiveness of the ``Surgical Video Inpainting'' module, we created a variant ``Diffusion+Deform3DGS'' by removing the mask processing from Deform3DGS and using the inpainting module as prior information. Compared with this baseline, Diff2DGS achieves better PSNR, SSIM, and RMSE on both datasets, indicating that the improvement is not only from the diffusion-based preprocessing but also from the proposed deformable 2D Gaussian representation and depth-aware optimization.
In the ``w/o deformation'' variant, we remove the Learnable Deformation Model (LDM). The performance degradation confirms the importance of explicitly modeling tissue deformation.
The ``w/o inpainting'' variant also reduces image reconstruction quality, showing that instrument removal is helpful for occlusion-aware reconstruction.
More importantly, removing the depth loss, i.e., ``w/o $L_{\mathrm{depth}}$ ($\lambda=0$)'', leads to much larger RMSE, although PSNR and SSIM remain competitive. This demonstrates that image-space metrics alone do not guarantee accurate geometry.
Finally, compared with the fixed-weight setting ``w/o adaptive ($\lambda=1$)'', the proposed adaptive depth weight significantly reduces RMSE from 8.58 to 3.79 on StereoMIS and from 6.74 to 2.51 on EndoNeRF, while also improving PSNR and SSIM. These results validate the effectiveness of the proposed adaptive depth loss weighting strategy.

\begin{figure}[htbp]
  \centering
  \includegraphics[width=1\columnwidth]{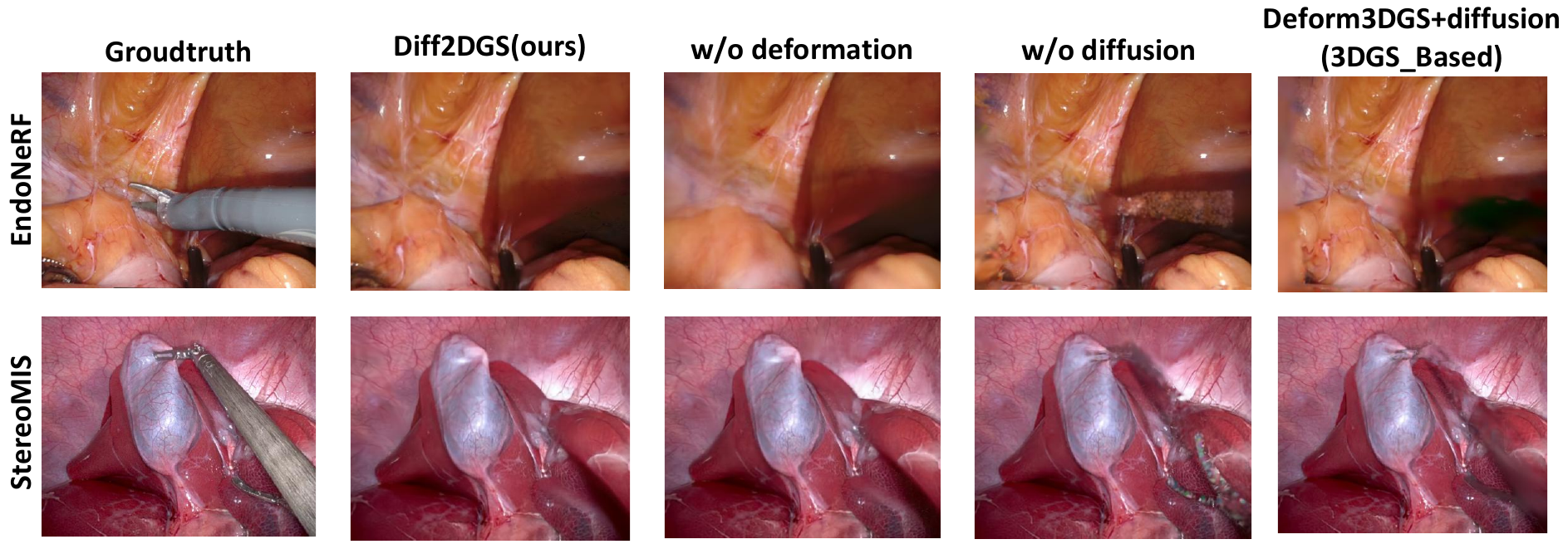}
  \vspace{-0.3cm}
  \caption{Ablation experiment on EndoNeRF and StereoMIS.}
  
  \label{fig:ablation}
\end{figure}

Fig.~\ref{fig:ablation} visualizes the ablation study on EndoNeRF and StereoMIS. Compared to the full model (Col.~2), removing the Learnable Deformation Model (LDM) (Col.~3) leads to significant detail loss and blurring, while omitting the inpainting module (Col.~4) results in severe artifacts, particularly in regions occluded by surgical instruments. Furthermore, comparing our 2DGS-based approach with the inpainting-enhanced Deform3DGS (Col.~5) demonstrates that while inpainting effectively mitigates occlusion artifacts for both approaches, our method yields more realistic textures and consistent geometric details in the reconstructed areas.

\vspace{-0.3cm}
\subsection{ {Error analysis of RAFT}}
Since neither EndoNeRF nor StereoMIS provides ground truth depth for occluded regions, and StereoMIS depth maps are derived from RAFT~\cite{raft}, we follow prior work and use RAFT to estimate depth for inpainted regions. To validate the reliability of RAFT as a reference, we conduct a quantitative error analysis on the SCARED dataset~\cite{scared}, which provides high-quality ground truth depth from structured light scanning.

We compare RAFT against two other state-of-the-art stereo matching methods: DLNR~\cite{dlnr} and STTR~\cite{sttr}. 
The evaluation is performed on the SCARED keyframes, and the results are summarized in Table~\ref{tab:scared_keyframes_comparison}. 
As shown, RAFT achieves the best performance in End-Point-Error (EPE) and Mean Absolute Error (MAE), with an MAE of approximately 1.01~mm, while its 3 px error, defined as the percentage of pixels whose stereo matching error exceeds 3 pixels, is also comparable to the best-performing method.
This demonstrates its high accuracy for surgical scene reconstruction and justifies our choice of RAFT as a reliable depth reference for evaluating the quality of recovered regions.

\begin{table}[htbp]
\vspace{-0.2cm}
  \centering
  \caption{Error analysis of RAFT on SCARED dataset.}
  \label{tab:scared_keyframes_comparison}
  \begin{tabular}{lccc}
    \toprule
    Method & EPE (px) $\downarrow$ & 3 px (\%) $\downarrow$ & MAE (mm) $\downarrow$ \\
    \midrule
    DLNR~\cite{dlnr}          & 1.45 & \textbf{4.12} & 1.32 \\
    STTR~\cite{sttr}      & 6.03 & 9.52 & 11.31 \\
    RAFT~\cite{raft} & \textbf{1.16} & 4.59 & \textbf{1.01} \\
    \bottomrule
  \end{tabular}
  \vspace{-0.3cm}
\end{table}

\section{Conclusion}
In this paper, we propose Diff2DGS for reliable reconstruction of occluded surgical scenes by combining diffusion-based tool inpainting, deformable 2D Gaussian Splatting, and adaptive depth weight. Experiments on EndoNeRF, StereoMIS, and SCARED show that Diff2DGS improves both appearance quality and geometric reliability while maintaining real-time rendering speed. The results further demonstrate that image similarity metrics alone are insufficient for evaluating surgical 3D reconstruction. Future work will extend the framework to scenarios with stronger camera motion.

\end{document}